\documentclass[prb,amsmath,twocolumn,superscriptaddress]{revtex4-1}
\usepackage{times}
\usepackage{url}
\usepackage{latexsym}
\usepackage{xcolor}
\usepackage{graphicx}
\usepackage{verbatim}
\usepackage{textcomp}

\begin{document}

\title{Discovering key topics from short, real-world medical inquiries via natural language processing and unsupervised learning}

\author{A. Ziletti}
\email{angelo.ziletti@bayer.com} 
\affiliation{Department of Decision Science \& Advanced Analytics, Bayer AG, 13353, Berlin, Germany}

\author{C. Berns}
\affiliation{Department of Data Science \& Data Engineering, areto consulting gmbh, 51063, Cologne, Germany}

\author{O. Treichel}
\affiliation{Department of Product Platforms, Bayer AG, 13353, Berlin, Germany}

\author{T. Weber}
\affiliation{Department of Product Platforms, Bayer AG, 13353, Berlin, Germany}

\author{J. Liang}
\affiliation{Department of Medical Information, Bayer AG, 13353, Berlin, Germany}

\author{S. Kammerath}
\affiliation{Department of Medical Affairs \& Pharmacovigilance, Bayer AG, 13353, Berlin, Germany}

\author{M. Schwaerzler}
\affiliation{Department of Decision Science \& Advanced Analytics, Bayer AG, 13353, Berlin, Germany}

\author{J. Virayah}
\affiliation{Department of Medical Affairs \& Pharmacovigilance Digital Transformation, Bayer AG, 13353, Berlin, Germany}

\author{D. Ruau}
\affiliation{Department of Decision Science \& Advanced Analytics, Bayer AG, 13353, Berlin, Germany}

\author{X. Ma}
\affiliation{Department of Integrated Evidence Generation \& Business Excellence, Bayer AG, 13353, Berlin, Germany}

\author{A. Mattern}
\affiliation{Department of Medical Information, Bayer AG, 13353, Berlin, Germany}

\begin{abstract} %141 words
% the motivation
Millions of unsolicited medical inquiries are received by pharmaceutical companies every year. 
It has been hypothesized that these inquiries represent a treasure trove of information, potentially giving insight into matters regarding medicinal products and the associated medical treatments. 
% the challenge
However, due to the large volume and specialized nature of the inquiries, it is difficult to perform timely, recurrent, and comprehensive analyses.
% the solution
Here, we propose a machine learning approach based on natural language processing and unsupervised learning to automatically discover key topics in real-world medical inquiries from customers. This approach does not require ontologies nor annotations. 
% the results
The discovered topics are meaningful and medically relevant, as judged by medical information specialists, thus demonstrating that unsolicited medical inquiries are a source of valuable customer insights.
% the implications and outlook
Our work paves the way for the machine-learning-driven analysis of medical inquiries in the pharmaceutical industry, which ultimately aims at improving patient care.
\end{abstract}

\maketitle
 
\section{Introduction}
\begin{figure*}[htb]
% trim is left bottom right top
\includegraphics[trim={0cm 0cm 0cm 0cm}, clip, width=\textwidth]{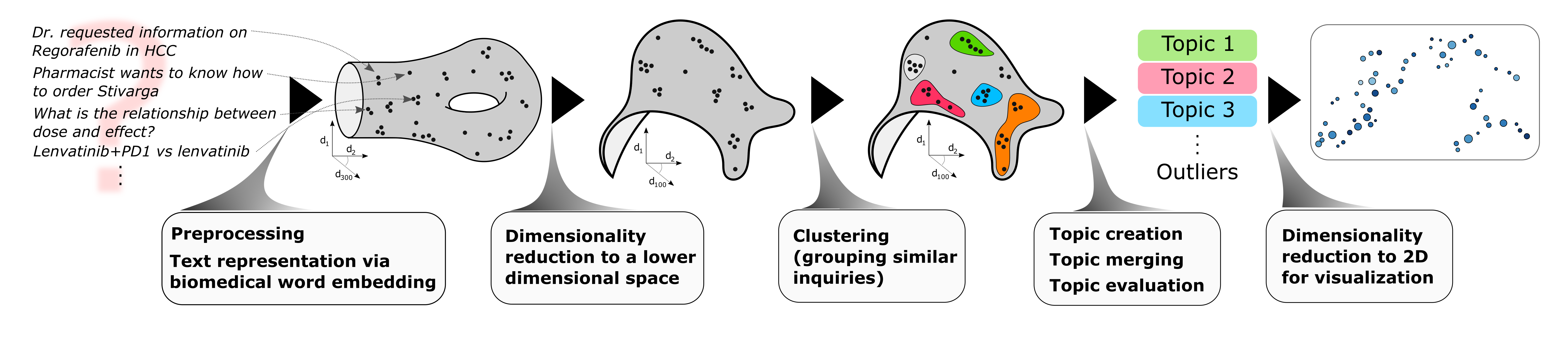}
\caption{From medical inquiries to medical topics via natural language processing and machine learning.}
 \label{fig:topic-modelling-chart}
\end{figure*}
\label{intro}
Every day pharmaceutical companies receive numerous medical inquiries related to their products from patients, healthcare professionals, research institutes, or public authorities from a variety of sources (\emph{e.g.} websites, e-mail, phone, social media channels, company personnel, telefax). 
These medical inquiries may relate to drug-drug-interactions, availability of products, side effects of pharmaceuticals, clinical trial information, product quality issues, comparison with competitor products, storage conditions, dosing regimen, and the like. 
On the one hand, a single medical inquiry is simply a question of a given person searching for a specific information related to a medicinal product. On the other hand, a plurality of medical inquiries from different persons may provide useful insight into matters related to medicinal products and associated medical treatments.
Examples of these insights could be early detection of product quality or supply chain issues, anticipation of treatment trends and market events, improvement of educational material and standard answers/frequently asked question coverage, potential changes in treatment pattern, or even suggestions on new possible indications to investigate.
From a strategic perspective, this information could enable organizations to make better decisions, drive organization results, and more broadly create benefits for the healthcare community. 

% transition paragraph - machine learning can help
However, obtaining high-level general insights is a complicated task since pharmaceutical companies receive  copius amounts of medical inquiries every year. Machine learning and natural language processing represent a promising route to automatically extract insights from these large amounts of unstructured (and noisy) medical text.
%
%
% text mining in general and in the biomedical domain
Natural language processing and text mining techniques have been widely used in the medical domain\cite{mehdi2017,luque-2019}, with particular emphasis on electronic health records\cite{sun-2017,kormilitzin-2020,mascio-2020,landi-2020}. 
In particular, deep learning has been successfully applied to medical text, with the overwhelming majority of works in supervised learning, or representation learning (in a supervised or self-supervised setting) to learn specialized word vector representations (\emph{i.e.} word embeddings)\cite{neumann-2019,beltagy-2019,alsentzer-2019,weng-2020,wu-2020}.
%
%There is little work however on unsupervised learning from unstructured medical text. 
Conversely, the literature on unsupervised learning for medical text is scarce despite the bulk of real-world medical text being unstructured, without any labels or annotations.
Unsupervised learning from unstructured medical text is mainly limited to the development of topic models based on latent Dirichlet allocation (LDA)\cite{blei-2003}. Examples of applications in the medical domain are clinical event identification in brain cancer patients from clinical reports\cite{arnold-2012}, modeling diseases\cite{pivovarov-2015} and predicting clinical order patterns\cite{chen-2017} from electronic health records, or detecting cases of noncompliance to drug treatment from patient forums\cite{redhouane-2018}.
Only recently, word embeddings and unsupervised learning techniques have been combined to analyze unstructured medical text to study the concept of diseases\cite{shah-2018}, medical product reviews\cite{karim-2020}, or to extract informative sentences for text summarization.\cite{moradi-2019}

% real-world corpus of medical inquiries and its challenges
In this work, we combine biomedical word embeddings and unsupervised learning to discover topics from real-world medical inquiries received by Bayer\texttrademark.
A real-world corpus of medical inquiries presents numerous challenges. From an inquirer (\emph{e.g.} healthcare professional or patient) perspective, often the goal is to convey the information requested in as few words as possible to save time. This leads to an extensive use of acronyms, sentences with atypical syntactic structure, occasionally missing verb or subject, or inquiries comprising exclusively a single noun phrase.
Moreover, since medical inquiries come from different sources, it is common to find additional (not relevant) information related to the text source; examples are references to internal computer systems, form frames (\emph{i.e.} textual instructions) alongside with the actual form content, lot numbers, email headers and signatures, city names. 
%
% mixture of layman and medical language
The corpus contains a mixture of layman and medical language depending (mostly) on the inquirer being either a patient or a healthcare professional. Style and content of medical inquiries vary quite substantially according to which therapeutic areas (\emph{e.g.} cardiovascular vs oncology) a given medicinal product belongs to.

% add sentence to refer to the text representation
%as one can see from Fig.\ref{fig:inquiries-histogram},
As already mentioned, medical inquiries are short. More specifically, they comprise less than fifteen words in the vast majority of cases. 
Standard techniques for topic modelling based on LDA\cite{blei-2003} do not apply, since the main assumption - each document/text is a distribution over topics - clearly does not hold given that the text is short\cite{qiang-2019}. 
Approaches based on pseudo-documents\cite{mehrotra-2013} or using auxiliary information\cite{phan-2008,jin-2011} are also not suitable since no meaningful pseudo-document nor auxiliary information are available for medical inquiries.
Moreoever, these models aim to learn semantics (\emph{e.g.} meaning of words) directly from the corpus of interest. However, the recent success of pretrained embeddings\cite{peters-2018,devlin-2018} shows that it is beneficial to include semantics learned on a general (and thus orders of magnitude larger) corpus, thus providing semantic information difficult to obtain from smaller corpora. This is particularly important for limited data and short text settings.
To this end, there has been recently some work aimed at incorporating word embeddings into probabilistic models similar to LDA (Dirichlet multinomial mixture model\cite{yin-2014}) and that - contrary to LDA - satisfies the single topic assumption (\emph{i.e.} one document/text belong to only one topic)\cite{nguyen-2015,li-2016}. 
Even though these models include (some) semantic information in the topic model, it is not evident how to choose the required hyper-parameters, for example determining an appropriate threshold when filtering semantically related word pairs.\cite{li-2016} Concurrently to our work, document-level embeddings and hierarchical clustering have been combined to obtain topic vectors from news articles and a question-answer corpus.\cite{angelov-2020}

% summary
Here, we propose an approach based on specialized biomedical word embeddings and unsupervised learning to discover topics from short, unstructured, real-world medical inquiries.
This approach - schematically depicted in Fig.\ref{fig:topic-modelling-chart} - is then used to discovery topics in medical inquiries received by Bayer\texttrademark\ Medical Information regarding the oncology medicinal product Stivarga\texttrademark.

\section{Results}

\subsection{Machine learning approach to discover topics in medical inquiries}
\label{sec:topic-discovery-algorithm}

\subsubsection{Text representation}
\label{sec:text-representation} 
\begin{table*}[htb]
\centering 
\begin{tabular}{l   l  l } 
\hline\hline 
Probe word &  Most similar words (standard embedding)  &  Most similar words (biomedical embedding) \\
\hline 
\emph{leukemia} & cancer (0.68), cancers (0.65), tumor (0.65), & leukaemia (0.97), leukemias (0.88), lymphoblastic (0.80), \\
& tumors (0.64), chemotherapy (0.63), marrow (0.63), & myelomonocytic (0.80), myelogenous (0.80), myeloid (0.80), \\
& prognosis (0.61), malignant (0.61), anemia (0.60), & promyelocytic (0.73), leukaemic (0.73), leukemic (0.72), \\
& diagnosed (0.60), pancreatic (0.59), ovarian (0.59) & blastic (0.67), blasts (0.67), therapy-related (0.66) \\
 \hline 
\emph{blood} & urine (0.63), bleeding (0.62), liver (0.61), & hematocrit (0.60), haematocrit (0.59), whole-blood (0.58), \\
& bloodstream (0.59), glucose (0.59), kidney (0.58), &  arterial (0.57), pressure (0.55),  heparinized (0.54),  \\
& heart (0.58), kidneys (0.57), cholesterol (0.57), &  oncotic (0.53), hemoglobin (0.53), haemoglobin (0.52), \\
& stomach (0.56), saliva (0.56), disease (0.56) & venous (0.52),  peripheral (0.52), venipuncture (0.51)\\ 
\hline 
\emph{carcinoma} & tumors (0.78), tumor (0.76), malignant (0.75), & carcinomas (0.90), adenocarcinoma (0.88), adenocarcinomas (0.79), \\
& cancers (0.74), ovarian (0.71), pancreatic (0.69), & squamous (0.76), well-differentiated (0.70), metastasizing (0.68),\\
& lesions (0.67), cancer (0.66), prognosis (0.66), & urothelial (0.68), tumours (0.68), cancers (0.68), \\
& lung (0.65), prostate (0.64), leukemia (0.60) & cancer (0.68), non-metastatic (0.67), tumors (0.66)\\
\hline\hline 
\end{tabular}
\label{table:most-similar-word}
\caption{\small Illustrative comparison between standard and biomedical word embeddings. The most similar words to the probe words \emph{blood}, \emph{carcinoma}, and \emph{leukemia} are shown for a standard and a biomedical word embedding. Values in parenthesis indicate the similarity with the corresponding probe word (maximum similarity is 1). The biomedical embedding model returns more specific and more medically relevant terms. The standard and biomedical embedding models are spaCy \emph{en\_core\_web\_lg} and scispaCy \emph{en\_core\_sci\_lg}, respectively.} 
\end{table*}

One of the main challenges of topic discovery in short text is sparseness: it is not possible to extract semantic information from word co-occurrences because words rarely appear together since the text is short. 
In our case, the sparseness problem is exacerbated by two following aspects.
First, the amount of data available is limited: most medicinal products receive less than 4,000 medical inquiries yearly.
Second, medical inquiries are sent by patients as well as healthcare professionals (\emph{e.g.} physicians, pharmacists, nurses): this leads to inquiries with widely different writing styles, containing a mixture of common and specialized medical text.
The sparsity problem can be tackled by leveraging word embedding models trained on large corpora; these embeddings have been shown to learn semantic similarities directly from data, even for specialized biomedical text \cite{neumann-2019,beltagy-2019,lee-2019,alsentzer-2019}. 
Specifically, we use the scispaCy word embedding model\cite{neumann-2019}, which was trained on a large corpus containing scientific abstracts from medical literature (PubMed) as well as web pages (OntoNotes 5.0 corpus\cite{pradhan-2013}).
This assorted training corpus enables the model to treat specialized medical terminology and layman terms on the same footing, so that medical topics are discovered regardless of the writing style.
%

% out-of-vocabulary words - cons of word2vec, alternatives
One of the main disadvantages of word vector (word2vec) models - like the (scispaCy) model used in this work - is their inability to handle out-of-vocabulary (oov) words: if a word appearing in the text is not included in the model vocabulary, it is effectively skipped from the analysis (\emph{i.e.} a vector of all zeros is assigned to it).
To tackle this issue, several models have been proposed, initially based on chargram level embeddings (FastText\cite{bojanowski-2016}), and more recently contextual embeddings based on character (ELMO\cite{peters-2018}), or byte pair encoding \cite{sennrich-2016} representations (BERT\cite{devlin-2018}).
Even though other advancements - namely word polysemy handling and the use of attention\cite{vaswani-2017} - were arguably the decisive factors, improvements in oov word handling also contributed in making ELMO and BERT the de-facto gold standard for natural language processing, at least for supervised learning tasks.

% why we use word2vec instead of ELMO (based onf characters) or BERT (based on byte-pair encoding)
Even though the use of contextual word embeddings is generally beneficial and can be readily incorporated in our approach (simply substituting the word representation), we notice that - given the large amount of noise present and the purely unsupervised setting - a word2vec model is actually advantageous for the task of extracting medical topics from real-word medical inquiries.
Indeed, using a model with a limited yet comprehensive vocabulary (the scispaCy model used in this work includes 600k word vectors) constitutes a principled, data-driven, efficient, and effective way to filter relevant information from the noise present in the corpus. This filtering is principled and data driven because the words (and vectors) included in the model vocabulary are automatically determined in the scispaCy training procedure by optimizing the performance on biomedical text benchmarks\cite{neumann-2019}.  
This also leads to harmonization of the medical inquiry corpus by eliminating both non-relevant region-specific terms, and noise introduced by machine translation (words or expressions are sometimes not translated but simply copied still in the original language \cite{knowles-2018}).
Clearly, in this context it is of paramount importance to use specialized biomedical embeddings so that the word2vec model has a comprehensive knowledge of medical terms despite its relatively limited vocabulary. 

Table \ref{table:most-similar-word} presents a qualitative comparison of a standard embedding (\emph{en\_core\_web\_lg}, trained on the Common Crawl) and a specialized biomedical embedding (scispaCy \emph{en\_core\_sci\_lg}, trained also on PubMed). Specifically, for a given probe word (\emph{i.e.} \emph{leukemia}, \emph{carcinoma}, \emph{blood}), the words most semantically similar to it - measured by the cosine similarity between word vectors - are retrieved, together with their similarity with the probe word (shown in parenthesis, 1.0 being the highest possible similarity). It is evident that the biomedical embedding returns much more relevant and medically specific terms. For instance, given the probe word \emph{leukemia}, the standard embedding returns generic terms like \emph{cancer}, \emph{tumor}, \emph{chemotherapy} which are broadly related to oncology, but not necessarily to leukemia. In contrast, the biomedical embedding returns more specialized (and medically relevant) terms like \emph{lymphoblastic}, \emph{myelomonocytic}, \emph{myelogenous}, \emph{myeloid}, \emph{promyelocytic}: acute lymphoblastic, chronic myelomonocytic, chronic myelogenous, adult acute myeloid, and acute promyelocytic are all types of leukemia.

\subsubsection{Clustering similar medical inquiries via hierarchical clustering}
\label{sec:clustering}

We have shown in the previous section that word embeddings provide a natural way to include semantic information (\emph{i.e.} meaning of individual words) in the modeling. 
Medical inquiries comprise multiple words, and therefore a semantic representation for each inquiry needs to be computed from the word-level embeddings. We accomplish this by simply averaging the embeddings of the words belonging to the inquiry, thus obtaining one vector for each inquiry.
Since these vectors capture semantic information, medical inquiries bearing similar meaning are mapped to nearby vectors in the high-dimensional embedding space. To group similar inquiries, clustering is performed in this embedding space, and for each medicinal product separately. 

Usually, it is not conducive to define an appropriate number of clusters \emph{a priori}. A reasonable number of clusters depends on various interdependent factors: number of incoming inquiries, therapeutic area of the medicine, time frame of the analysis, and intrinsic amount of information (\emph{i.e.} variety of the medical inquiries).
For a given medicinal product, typically a handful of frequently asked questions covers a large volume of inquiries, accompanied by numerous low-volume and less cohesive inquiry clusters. These low-volume clusters often contain valuable information, which might not even be known to medical experts: their low volume makes it difficult to detect them via manual inspection.
To perform clustering in the embedding space, we use the hierarchical, density based clustering algorithm HDBSCAN \cite{campello-2013,mcinnes-2017,mcinnes-2017b}.
As customary in unsupervised learning tasks, one needs to provide some information on the desired granularity, \emph{i.e.} how fine or coarse the clustering should be.
In HDSBCAN, this is accomplished by specifying a single, intuitive hyper-parameter (\emph{min\_cluster\_size}). In our case, the objective is to obtain approximately 100 clusters so that the results can be easily analyzed by medical experts. 
Thus, the main factor in defining \emph{min\_cluster\_size} is the number of inquiries for a given medicinal product: the larger the medical inquiry volume, the larger the parameter \emph{min\_cluster\_size}.
Note that \emph{min\_cluster\_size} is not a strict controller of cluster size (and thus how many clusters should be formed), but rather a guidance provided to the algorithm regarding the desired clustering granularity. It is also possible to combine different  \emph{min\_cluster\_size} for the same dataset, \emph{i.e.} using a finer granularity for more recent inquiries, thus enabling the discovery of new topics when only few inquiries are received, at a price however of an increase in noise given the low data volume. Moreover, \emph{min\_cluster\_size}  is very slowly varying with data (medical inquiry) volume, which facilitate its determination (see Methods).

% UMAP before clustering
In practice, a non-linear dimensionality reduction is applied to lower the dimensionality of the text representation before clustering is performed. We utilize the UMAP algorithm\cite{mcinnes-2018,mcinnes-2018b} because of its firm mathematical foundations from manifold learning and fuzzy topology, ability to meaningfully project to any number of dimensions (not only two or three like t-SNE\cite{vanDerMaaten-2008}), and computational efficiency.
Reducing the dimensionality considerably improves the clustering computational performance, greatly easing model deployment to production, especially for products with more than 5,000 inquiries. 
The dimensionality reduction can in principle be omitted, especially for smaller datasets.
At the end of this step, for each medicinal product a set of clusters is returned, each containing a collection of medical inquiries. A given medical inquiry is associated to one topic only, in accordance with the single topic assumption.

% topic naming
In order to convey the cluster content to users, a name (or headline) needs to be determined for each cluster.
To this end, the top-five most recurring words for each cluster are concatenated, provided that they appear in at least 20\% of the inquiries belonging to that cluster; this frequency threshold is set to avoid to include in the topic name words that appear very infrequently but are still in the top-five words. Thus, if a word does not fulfill the frequency requirement, it is not included in the topic name (resulting in topic names with less than five words). By such naming (topic creation), the clusters are represented by a set of words, which summarize their semantic content.
%
% merge topics based on words
Finally, topics with similar names are merged in order to limit the number of topics to be presented to medical experts (see Methods).
%
% final outcome
After the topics are merged, new topic names are generated according to the procedure outlined above. The final result is a list of topics defined by a given name, each containing a set of similar medical inquiries. The list of discovered topics is then outputted, and presented to medical experts.

% topic map for visualization of topic similarity
Since the goal is to extract as much knowledge as possible from incoming medical inquiries, a relatively large number of topics (typically around 100) is returned to medical experts for each medicinal product.
To facilitate topic exploration and analysis, topics are visualized on a map that reflects the similarity between topics (Fig. \ref{fig:topic-map-stivarga}a): topics close to each other in this map are semantically similar.
To obtain this semantic map, first topic vectors are computed by averaging the text representation of all inquiries belonging to a given topic; then, a dimensionality reduction to two dimensions via UMAP is performed. 

\subsubsection{Topic evaluation: topic semantic compactness and name saliency}
Once topics are discovered, it is desirable to provide medical experts with information regarding the quality of a given topic. To this end, a score is calculated for each topic. 
Since no available metric applies to the case of medical topic discovery (see Methods), we introduce two new quantities to evaluate topics discovered in an unsupervised setting. 
These quantities - which we term topic semantic compactness and name saliency - fully leverage semantic similarity at the sentence (medical inquiry) level; they are also intuitive, computationally efficient, and intrinsic, \emph{i.e.} do not require any gold labels.

One way of quantifying the quality of a discovered topic is to determine how similar the inquiries grouped in a topic are: intuitively, the more similar the inquiries in a topic are, the higher the quality of this topic.
As described in Sec. \ref{sec:text-representation}-\ref{sec:clustering}, we maps inquiries (via word2vec and averaging of word vectors) in a semantic space where clustering is performed. From a geometrical point of view, topic quality can be estimated by calculating the similarity among medical inquiries within a given topic in this semantic space; those similarities are then summed to evaluate the semantic compactness of a topic. 
The topic semantic compactness is between zero and one, one indicating the highest possible compactness.

The topic name is one of the main information shown to the users to summarize the semantic content of a discovered medical topic. It is therefore of interest to quantify how representative the name is for a given medical topic. This is tackled by answering the following question: how similar is the name with the inquiries grouped in the topic it represents?
This is quantified by the name salience, a score between zero and one, the higher the value, the more salient the name w.r.t. the topic its represents. Topic quality is then obtained by averaging semantic compactness and title saliency for each topic. Additional information on topic evaluation can be found in Methods.

\subsection{A real-world example of topic discovery: the oncology product Stivarga\texttrademark}

\begin{figure}[thb]
% trim is left bottom right top
\includegraphics[trim={0cm 0cm 0cm 0cm}, clip, width=0.5\textwidth]{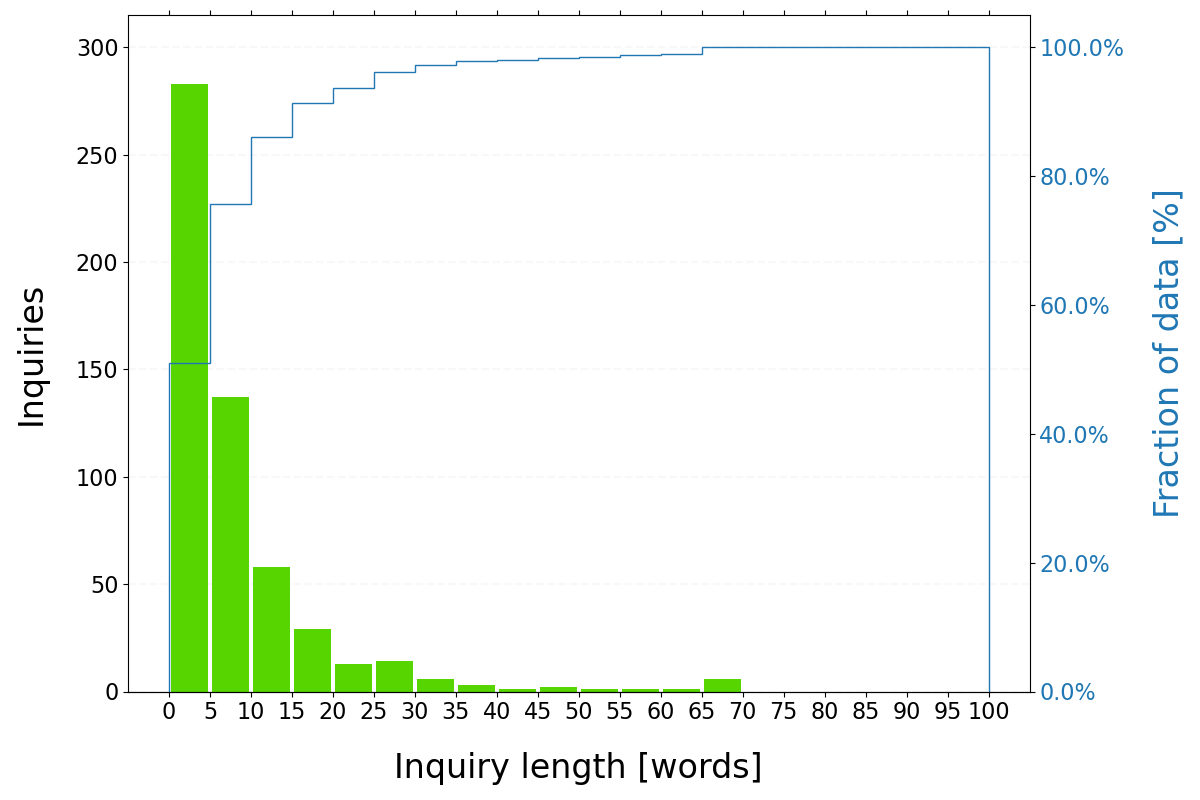}
\caption{Length distribution of medical inquiries on Stivarga\texttrademark. The large majority of the inquiries ($\sim$90~\%) contains less than fifteen words after preprocessing.}
 \label{fig:inquiries-histogram}
\end{figure}

\begin{figure*}[h!tb]
% trim is left bottom right top
\includegraphics[trim={0cm 0cm 0cm 0cm}, clip, width=\textwidth]{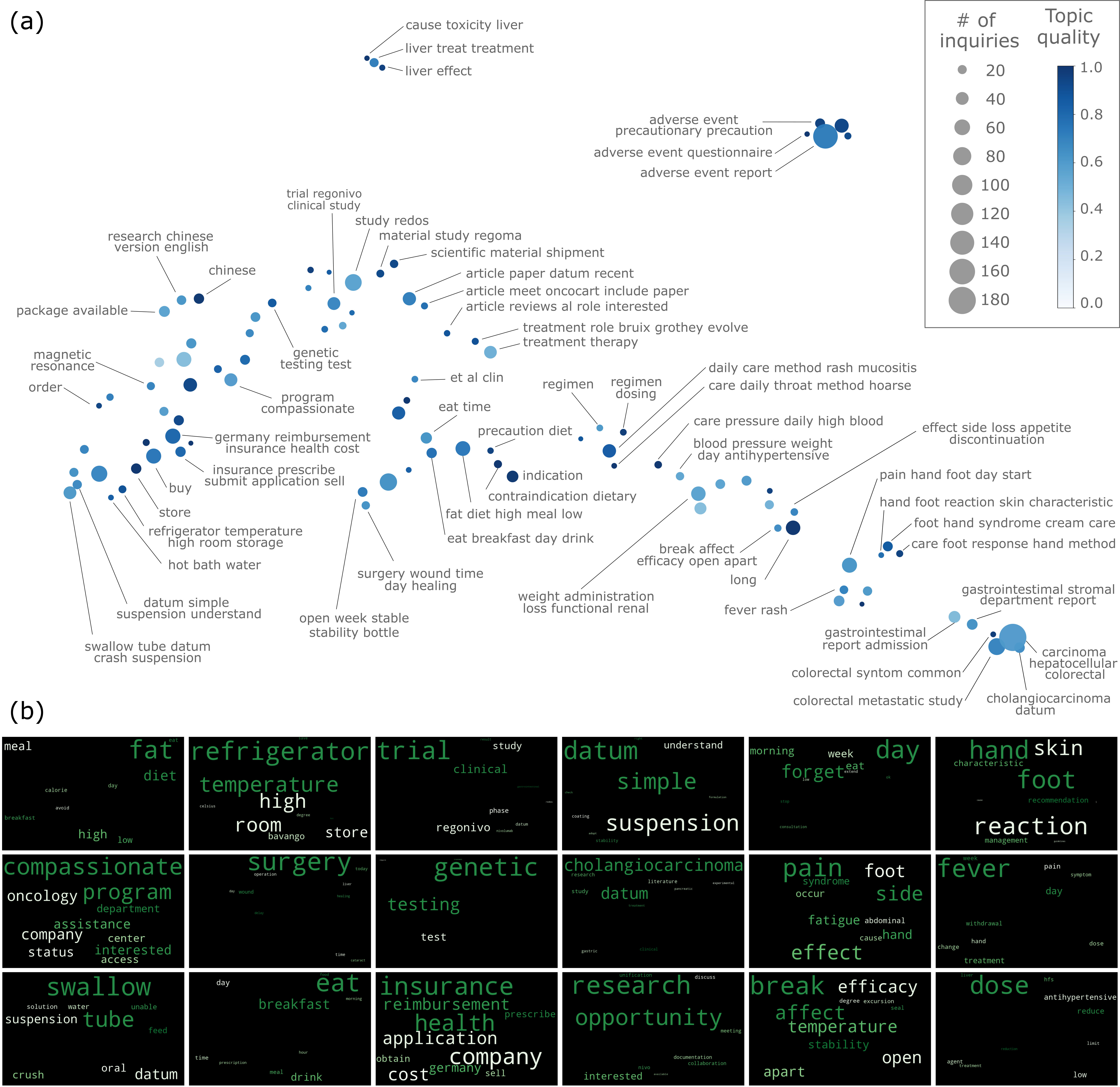}
\caption{Topic discovery for medical inquiries on Stivarga\texttrademark\. (a) Topic semantic map. Axis are the non-linear dimensionality components and thus are not shown. (b) Wordclouds for selected topics.}
 \label{fig:topic-map-stivarga}
\end{figure*}

% introduction on Stivarga
As a real-world example of topic discovery, we present the results for medical inquiries on the oncology product Stivarga\texttrademark. Stivarga\texttrademark\ is an oral multikinase inhibitor which inhibits various signal pathways responsible for tumor growth. 

% more details
In this work, all unsolicited medical inquiries received by Bayer\texttrademark\ worldwide in the time frame July 2019-June 2020 are considered. All non-English inquiries are translated to English using machine translation. These inquiries are then pre-processed: acronyms are resolved; non-informative phrases, words or patterns are removed; text is tokenized and lemmatized. Additional details are provided in Methods. Then, the topic discovery algorithm introduced in Sec. \ref{sec:topic-discovery-algorithm} is applied.

% results introduction
The semantic map with the discovered topics is shown in Fig. \ref{fig:topic-map-stivarga}a. 
These topics span a relatively large variety of themes, ranging from interactions with food and adverse drug reactions to purchase costs and literature requests.
The topics are judged as meaningful and medically relevant by medical information specialists, on the basis of their expert knowledge of the medicinal product.

Topics are also specific: the unsupervised learning approach allows information to emerge directly from the data, without recurring to predefined lists of keywords or classes, as required when using ontologies or supervised learning.
An example of a very specialized topic for inquiries on scientific literature is \emph{treatment role bruix grothey evolve}: 12 requests related to the review article on the treatment of advanced cancer with Regorafenib (active ingredient of Stivarga\texttrademark) published on February 2020 \cite{grothey-2020}. 
Other examples are the five topics \emph{fat diet high meal low}, \emph{eat breakfast day drink}, \emph{precaution diet}, \emph{eat time}, \emph{contraindication diet}. Even though all these topics relate to nutrition, they are addressing different aspects. It is quite advantageous that they are identified as distinct since medical recommendations will likely differ across these five topics. 

Thanks to the inclusion of semantics via word embedding, the algorithm is able to group together inquiries having similar meaning, even though the actual words in them are distinct. For instance, the topic \emph{pain side effect foot fatigue} comprises 21 inquiries on medical issues (which may or may not be related with the medicine), in which the following words appear: \emph{pain} (seven times), \emph{side effect} (six times) \emph{nausea} (three times), \emph{fatigue} (five times), \emph{dysphonia} (two times). The algorithm is able to cluster these inquiries together because similar inquiries are mapped close to each other in the high dimensional semantic space where clustering is performed. This is corroborated by the relatively high similarity score between the terms appearing in these inquiries (pain-nausea: 0.66, nausea-fatigue:0.61, pain-fatigue:0.71, dysphonia-pain:0.55, dysphonia-fatigue:0.49), scores much higher than zero, zero being the score expected for unrelated terms.(\emph{cf.} pain-day:0.05, nausea-sun:0.08).
Conversely, if there is a moderate number of inquiries on a specific medical matter, the algorithm is generally able to detect that signal, as in the case of mucositis and hoarse in the two topics \emph{daily care method rash mucositis}, and \emph{daily care method rash hoarse}.

% the name is not all information
As shown in Fig. \ref{fig:topic-map-stivarga}a, the automatically generated topic names (see Sec.\ref{sec:clustering}) provide a reasonably good insight into the semantic content of their respective topics. However, one needs to be mindful that the topic might - and usually will - contain additional information of relevance. To convey this information in a simple yet effective way to the users, wordclouds are generated for each topic; examples of wordcloud are shown in Fig. \ref{fig:topic-map-stivarga}b. 
%
% example of compassionate program
For example, in the wordcloud of topic \emph{compassionate program} (Fig. \ref{fig:topic-map-stivarga}b, 1st column-2nd row), concepts not included in the topic name (\emph{e.g.} \emph{assistance}, \emph{interested}, \emph{access}, \emph{status}) appear, thus giving further insight into the topic content.
%
% for some topic, to understand them one needs to inspect the inquiries
In some cases, even the wordcloud might not convey the topic meaning: users will then resort to manually inspect the inquiries belonging to the topic.
For instance, the content of topic \emph{chinese} is not clear, neither from the topic name nor from the wordcloud; however, inspection of the actual inquiries quickly reveals that they refer to the interaction between Chinese medicine and Stivarga\texttrademark (the word \emph{medicine} does not appear since it is a stopword). Another examples are \emph{al et clin} and \emph{long}, which group together requests for scientific articles and product durability, respectively.

% topic quality
Topic quality provides a useful guidance when exploring topics. If topic quality is close to one, medical inquiries in that topic are all very similar, and the topic name is expected to summarize the topic content well. Conversely, topics with low quality will contain inquiries that might differ quite substantially, yet are similar enough to be clustered together by the algorithm. In these cases, manual inspection of the underlying medical inquiries may be a good strategy. 
From Fig. \ref{fig:topic-map-stivarga}a, it appears that smaller topics tend to have higher topic scores, although no clear trend emerges.

% semantic map
Finally, in addition of having similar inquiries within topics, the model captures semantic similarities between topics. This is apparent from Fig. \ref{fig:topic-map-stivarga}a: similar topics tend to be close to each other in the semantic map.  
Even though this feature does not influence the topic discovered, from a user perspective it provides a clear advantage when exploring topics (\emph{e.g.} compared to reading them from as a simple list).

\section{Discussion}
%advantages
This study introduces an unsupervised machine learning approach to automatically discover topics from medical inquiries. 
After the initial (one-time) effort for preprocessing (\emph{e.g.} abbreviation definition, stopword refinement) and hyper-parameters determination, the algorithm runs without requiring any human intervention, discovering key topics as medical inquiries are received. 
Topics can be discovered even if only a small number of inquiries is present, and are generally specific, thus enabling targeted, informed decisions by medical experts. 
Being completely unsupervised, the algorithm can discover topics that were neither known nor expected in advance, topics which often are the most valuable.
This is in stark contrast with ontology or supervised based approaches, where topics need to be defined \emph{a priori} (as collections of keywords or classes), and incoming text can be associated only to these predefined lists of topics, thus hindering the discovery of \emph{a priori} unknown topics.
The machine learning approach introduced here does not use ontologies (which are costly and hard to build, validate, maintain, and difficult to apply when layman and specialized medical terms are combined), and instead it incorporates domain knowledge via specialized biomedical word embeddings.
This allows to readily apply the topic discovery algorithm to different medicinal products, without the burden of having to develop specialized ontologies for each product or therapeutic area. Indeed, the algorithm is periodically analyzing medical inquiries for a total of sixteen Bayer\texttrademark\ medicinal products, encompassing cardiology, oncology, gynecology, hematology, and ophthalmology.

% disadvantages
Our approach has several limitations.
First, it can happen that a small fraction of inquiries associated to a given topic are actually extraneous to it, especially for semantically broad topics. 
This is because - due to the noise present in this real-world dataset - the soft clustering HDBSCAN algorithm must be applied with a low probability threshold for cluster assignment to avoid the majority of inquiries being considered as outliers (see Methods). 
Second, even though the topic names are generally quite informative, a medical expert needs to read the actual inquiries to fully grasp the topic meaning, especially if a decision will be made on the grounds of the discovered topics. This is however not burdensome because inspection is limited to the inquiries associated to a given topic (and not all inquiries).
Last, some discovered topics are judged by medical experts - based on their expert knowledge - so similar that they could have been merged in a single topic, but are considered distinct by the algorithm. In these cases, manual topic grouping might be required to determine the top topics by inquiry volumes. Still, these similar topics very often appear close to each other in the topic map.

% value despite the limitations
Despite these limitations, this study demonstrates that medical inquiries contain useful information, and that machine learning can extract this information in an automatic way, discovering topics that are judged by medical information specialists as meaningful and valuable. The hope is that this will stimulate mining of medical inquiries, and more generally the use of natural language processing and unsupervised learning in the medical industry.
Interesting future directions are the inclusion of \emph{a priori}  expert knowledge (\emph{e.g.} a list of expected topics) while at the same time maintaining the ability to discover new and previously unknown topics, and grouping topics in meta-topics though a clustering hierarchy.

\section{Methods}

\subsection{Preprocessing}
Since our dataset comprises real-word medical inquiries, preprocessing is a crucial step to limit the amount of noise in the corpus. 
% acronyms
The corpus contains numerous acronyms: a first step is thus acronym resolution, \emph{i.e.} substitute a given acronym with its extended form.
A dictionary for the most recurring acronyms ($\sim$40 per product) is compiled with the help of medical experts.
Acronym resolution is performed via a curated dictionary for two reasons. First, the data is too scarce and noisy to train a reliable, custom-made word embedding to learn the acronym meanings from the corpus. Second, in pretrained word embeddings typically there is no suitable representation for the acronym, or the acronym in our corpus is used to indicate something different than in natural language (it can even be company-specific). For example, in our corpus \emph{lode} does not refer to a vein of a metal, but stands for \emph{lack of product effect}. 
Regular expressions are then used to remove non-informative strings (\emph{e.g.} lot numbers, references to internal systems).
Next. text is split into sentences, tokenized and lemmatized using the scispaCy library \cite{neumann-2019} (which is built on top of spaCy). We disable the scispaCy parser; this gives a significant speed-up without affecting the topic discovery outcome.
Finally stopwords (\emph{i.e.} non informative words) are removed. In addition to standard English stopwords, there are non-standard stopwords which arise from the dataset being composed of medical inquiries \emph{e.g.} \emph{ask, request, email, inquiry, patient, doctor}, and product-dependent stopwords, typically the brand and chemical name of the medicinal product to which the inquiries refer to.
It is also the case that in the medical inquiry corpus single words bear value, but when combined they are no longer relevant for medical topic discovery. 
For example, the word \emph{years} and \emph{old} are generally of relevance, but if contiguous (\emph{years old}) they are no longer significant since this expression simply originates from a medical information specialists logging the age of the patient to which the inquiry refers to. 
Another example is the word \emph{morning}: when appearing alone it is of relevance, but when it is preceded by the word \emph{good} it loses its relevance since the expression \emph{good morning} does not bear any significance for medical topic discovery. We compile a short list of stop n-grams ($\sim$20) and remove them from the corpus.

\subsection{Text representation}
To represent medical inquiries, the scispaCy\cite{neumann-2019} word embedding model \emph{en\_core\_sci\_lg-0.2.5} is used.
No model re-training or fine-tuning is performed because of the small amount of data and the sparsity problem; since no labels are available, one would need to train a language model on noisy and short text instances which would likely lead the model to forget the semantics learned by the scispaCy model.
For each token, the (200-dimensional) scispaCy embedding vector is retrieved; the sentence representation is then obtained simply by calculating the arithmetic average of the vectors representing each token over all tokens belonging to a given sentence. 

\subsection{Out-of-vocabulary words}
Even though the overwhelming majority of out-of-vocabulary (oov) words are not of interest for medical topic discovery, a very small (but relevant) subset of important oov words would be missed if one were to simply use the word2vec model. We thus devise a strategy to overcome this, as described below.
For each product, the most recurring oov words are automatically detected; these words need to be included in the word2vec model so that they can be represented by a vector which accurately captures their meaning.
Training a new embedding to include these new terms is not a good approach given the sparseness problem described above. 
To overcome this, we combine a definition mapping and embedding strategy.
% definition mapping of out-of-vocabulary words
Specifically, first each of the relevant oov terms is manually mapped to a short definition; for example, the oov \emph{ReDOS} is mapped to \emph{dose optimization study} since ReDOS refers to a dose-optimisation phase 2 study on regorafenib \cite{bekaii-saab-2018}. 
% definition embedding
Then, using the text from these definitions, a meaningful vector representation for the oov words is obtained with the embedding strategy described above (scispaCy word embedding model and arithmetic average of word vectors, the word vectors averaged now being the words of the oov word definition).
This procedure has two main benefits.
First, it does not require any training data nor any training effort.
Second, it ensures by construction that the added word vectors are compatible with the word representation model in use.
Pharmaceutical product trade names are oov words of particular interest for medical topic discovery. 
%Indeed, being able to take into consideration drug trade names is of importance since there is a substantial amount of questions which mention for instance drug interactions. 
However, they are are generally not included in the scispaCy model. Thus, a slightly different procedure is used to ensure that all trade names appearing in medical inquiries are added to the model, regardless of them belonging to the most recurring oov words or not. 
Luckily, international non-proprietary names (INNs) of drugs are included. For instance, the oncology product trade name \emph{Stivarga}\texttrademark\ is not present, while its corresponding INN (\emph{regorafenib}) is. 
Thus, to automatically detect drug trade names we utilize the scispaCy named entity recognizer (NER) and the scispaCy UmlsEntityLinker as follows.  
First, the NER is used to extract entities from the text; then, for each entity, the UmlsEntityLinker performs a linking with the Unified Medical Language System (UMLS)\cite{bodenreider-2004} by searching within a knowledge base of approximately 2.7 million concepts via string overlap as described in Ref. \onlinecite{neumann-2019}.
To limit the number of false positive matches we increase the UmlsEntityLinker threshold to 0.85 from the default of 0.7.
For each entities that has been successfully linked to UMLS, several information regarding the identified concepts are returned by the UmlsEntityLinker: concept unique identifier (CUI), concept preferred name, concept definition, concept aliases, and concept type unique identifier (TUI). In particular, the latter defines to which semantic group the linked concept belongs to \cite{mccray-2001,bodenreider-2003}; an up-to-date list of semantic type mappings can be found at \cite{bodenreider-2020}. A TUI value of T121 indicates that the concept found is a \emph{Pharmacologic Substance}. Extracting the entities with TUI equal to T121 allows to automatically identify drug trade names. 
Each drug trade name is then mapped to the concept preferred name; if that is not present, the concept definition is used; if that is also not present, drug trade name is replaced by to the phrase \emph{pharmaceutical medication drug}. 
Once this mapping is performed, the same embedding strategy used for the other oov words is followed in order to obtain semantically meaningful word vector representations.

\subsection{Hierarchical clustering}
The HDBSCAN algorithm\cite{campello-2013,mcinnes-2017,mcinnes-2017b} starts by defining a mutual reachability distance based on a density estimation; the data is then represented as a weighted graph where vertices are data points and edges have weight equal to the mutual reachability distance between points. 
The minimum spanning tree is built, and converted to a hierarchy of connected components via a union-find data structure: starting from an initial cluster containing all points, the data is subsequently split at each level of the hierarchy according to the distance, ultimately returning as many clusters as data points when the threshold distance approaches zero. This cluster hierarchy is commonly depicted as dendogram.
To obtain a meaningful set of clusters, this hierarchy needs to be condensed. The crucial point is to discern - at any given split - if two new meaningful clusters are formed by splitting their parent cluster, or instead the parent cluster is simply loosing points (and in the latter case one wishes to keep the parent cluster undivided).
In HDBSCAN, this decision is governed by the minimum cluster size hyper-parameter (\emph{min\_cluster\_size}): a cluster split is accepted only if both newly formed clusters have at least \emph{min\_cluster\_size} points. The final clusters are then chosen from this set of condensed clusters by means of a measure of stability as defined by Ref. \onlinecite{campello-2013}.
%
%
% how we define the hyperparameters
The main factor in defining \emph{min\_cluster\_size} is the number of inquiries for a given product: we want to obtain (approximately) 100 clusters so that results can be easily analyzed by medical experts. 
It is important to point out that \emph{min\_cluster\_size} does not strictly specify the number of clusters that will be formed, but rather provides to the algorithm an indication regarding the desired granularity, as outlined above. In our case, \emph{min\_cluster\_size} ranges only between 5 and 10 depending on the number of inquiries. This small range of variation substantially facilitate the hyper-parameter search.
Moreover, we noticed that - for approximately the same amount of inquiries and same \emph{min\_cluster\_size} - the number of returned clusters increases with data variety, where data variety is qualitatively evaluated by manual inspection: for products with more diverse inquiries HDBSCAN tends to return a higher number of clusters, \emph{ceteris paribus}. % ceteris paribus means all things being equal
We utilize the leaf cluster selection method instead of the excess of mass algorithm because the former is known to return more homogeneous clusters \cite{mcinnes-2017}.

% we use the soft clustering
Due to the noise in the dataset, using the standard (hard) HDBSCAN clustering results in a large portion of the dataset (30-60\%) considered as outliers consistently across all products. 
To overcome this, we use the soft HDBSCAN clustering\cite{mcinnes-2017}, which returns - instead of a (hard) cluster assignment - the probability that an inquiry belongs to a given cluster. We then define a probability threshold under which a point is considered to be an outlier; for all other points above this threshold, we associate them to the cluster with the highest probability through an argmax operation. This probability threshold ranges between $10^{-3}$ and $10^{-2}$  and it is chosen such that approximately 10\% of the inquiries are classified as outliers.
As mentioned in the main text, for computational reasons, we project via UMAP to a lower dimensional space before clustering is performed. Specifically, we project to 100 dimensions for products with less than 15,000 inquiries, and to 20 dimensions for products with more than 15,000 inquiries.
Moreover, inquiries longer than 800 characters are also considered as outliers: this is because the text representation (average of word vectors) degrades for long sentences. These inquiries are gathered in the outlier cluster and made available to medical experts for manual inspection.

\subsection{Topic merging}
Given a topic, the vector representation for each word in the topic name is calculated; the topic name vector is then obtained by averaging the word vectors of the words present in the topic name. Topics are merged if their similarity - evaluated as cosine similarity between their topic name vectors - is larger than a threshold. Threshold values range between 0.8 and 0.95 depending on the medicinal product considered. 

\subsection{Topic evaluation: topic semantic compactness and name saliency}
The most popular topic evaluation metrics for topic modelling on long text are UCI \cite{newman-2010} and UMass \cite{mimno-2011}.
However, both UCI and UMass metrics are not good indicators for quality of topics in short text topic modelling due to the sparseness problem\cite{quan-2015}.
In Ref. \onlinecite{quan-2015}, a purity measure is introduced to evaluate short text topic modelling; however, it requires pairs of short and long documents (\emph{e.g.} abstract and corresponding full text article), and thus it is not applicable here because there is no long document associated to a given medical inquiry. Indeed, evaluation of short text topic modelling is an open research problem \cite{qiang-2019}. 
An additional challenge is the absence of labels. Performing annotations would require substantial manual effort by specialized medical professionals, and would be of limited use because one of the main goals is to discover previously unknown topics as new inquiries are received.
The absence of labels precludes the use of the metrics based on purity and normalized mutual information proposed in Ref. \onlinecite{rosenberg-2007}, \onlinecite{huang-2013}, \onlinecite{yin-2014}.

% distributional semantic
Ref. \onlinecite{aletras-2013} bring forward the valuable idea of using distributional semantic to evaluate topic coherence, exploiting the semantic similarity learned by word2vec models. Topic coherence is assessed by calculating the similarity among the top n-words of a given topic: semantically similar top n-words lead to higher topic coherence. If this might be in general desirable, in the case of discovering medical topics it is actually detrimental: interesting (and potentially previously unknown) topics are often characterized by top n-words which are not semantically similar. 
For example, a medical topic having as top 2-words \emph{rivaroxaban} (an anticoagulant medication) and \emph{glutine} is clearly relevant from a medical topic discovery standpoint. However, \emph{rivaroxaban} and \emph{glutine} are not semantically similar, and thus the metric proposed in Ref. \onlinecite{aletras-2013} would consider this as a low coherence (and thus low quality) topic, in stark contrast with human expert judgment.  
Analogous considerations apply to the indirect confirmation measures in Ref. \onlinecite{roeder-2015}: words emerging in novel topics would have rarely appeared before in a shared context.
For this reason, we introduce a new measure of topic compactness which takes into account the semantics of the inquiries, and does not require any labeled data.
Specifically, we compute the similarity of all inquiries belonging to a given topic with each other (excluding self-similarity), sum the elements of the resulting similarity matrix, and divide by the total number of elements in this matrix. The topic semantic compactness $\gamma^\alpha$ of topic $\alpha$ reads
\begin{equation}
\label{eq:topic-compactness}
\gamma^\alpha = \sum_{i=1}^{\vert C^\alpha \vert} \sum_{\scriptstyle j=1 \atop\scriptstyle i\ne j}^{\vert C^\alpha \vert} \frac{\mathcal{S}(q_i, q_j)}{\vert C^\alpha \vert (\vert C^\alpha \vert -1)}
\end{equation}
where $\vert C^\alpha \vert$ is the cardinality of topic $\alpha$ (how many inquiries are in topic $\alpha$), $q_i$ (and $q_j$) is the word vector representing inquiry $i$ ($j$), and $\mathcal{S}$ is a function quantifying the semantic similarity between inquiry $q_i$ and $q_j$, taking values between 0 and 1 ($\mathcal{S}=1$ when $q_i$ and $q_j$ are indentical, and $\mathcal{S}=0$ being the lowest possible similarity).
Given the chosen normalization factor (\emph{i.e.} the denominator in Eq.\ref{eq:topic-compactness}), $0 \le \gamma^\alpha \le 1$ and thus $\gamma^\alpha$ can be directly used as (a proxy for) topic quality score. 
The topic compactness maximum ($\gamma^\alpha=1$) is attained if and only if every sentence (after preprocessing) contains exactly the same words.
It is important to point out that $\gamma^\alpha$ automatically takes semantics into account: different but semantically similar medical inquiries would still have high similarity score, and thus would lead (as desired) to a high topic semantic compactness, despite these inquiries using different words to express similar content. 

% add here example of glutine
Contrary to Ref. \onlinecite{aletras-2013}, the topic semantic compactness $\gamma^\alpha$ introduced in Eq.\ref{eq:topic-compactness} does not artificially penalize novel topics just because they associate semantically different words appearing in the same inquiry.
To come back to the previous example, if numerous inquiries in a discovered topic contain the words \emph{rivaroxaban} and \emph{glutine}, the topic semantic compactness would be high (as desired), regardless from the fact that the top 2-words are not semantically similar since the similarity is evaluated at the inquiry level (by $\mathcal{S}(q_i, q_j)$ in Eq. \ref{eq:topic-compactness}).

It is also beneficial to evaluate how representative the topic name is for the topic it represents.
To this end, we calculate the name saliency $\tau^\alpha$ for medical topic $\alpha$ by calculating the similarity of the word vector representing the topic name with the word vectors representing the inquiries in the topic, sum these similarity values, and divide by the total number of inquiries in the topic. This reads
\begin{equation}
\label{eq:name-saliency}
\tau^\alpha = \sum_{i=1}^{\vert C^\alpha \vert} \frac{\mathcal{S}(t^\alpha, q_i)}{\vert C^\alpha \vert}
\end{equation}
where $\vert C^\alpha \vert$ is the cardinality of topic $\alpha$ (how many inquiries are in topic $\alpha$), $t^\alpha$ is the word vector representing the name of topic $\alpha$, and $q_i$ is the vector representing inquiry $i$. 
This returns a score ($ 0 \le \tau^\alpha \le 1$) which quantifies how representative (salient) the name is for the topic it represents.
As in the case of the topic semantic compactness, the name saliency $\tau^\alpha$ takes natively semantics (\emph{e.g.} synonyms) into account via $\mathcal{S}(t^\alpha, q_i)$ in Eq. \ref{eq:name-saliency}.
In both Eq. \ref{eq:topic-compactness} and Eq. \ref{eq:name-saliency}, the cosine similarity is used as similarity measure.

\section{Competing interests}
Financial support for the research was provided by Bayer AG.
The authors reports a
 patent application on \emph{Topic Modelling of Short Medical Inquiries} submitted on April 21st, 2020 (application number EP20170513.4). 

\bibliography{dimi-nlp-paper-npj-digital-medicine}
%\bibliography{try}
\bibliographystyle{ieeetr}{}

\section*{Author Contributions}
A.Z. led and thereby ideated and implemented the topic discovery algorithm, and is the main author the manuscript. M.S., C.B., D.R. provided valuable suggestions on the topic discovery algorithm. C.B., O.T., and T.W. designed and implemented the software architecture and data engineering pipeline for the algorithm deployment. T.W., J.V., J.L., S.K., X.M., A.M., D.R., and M.S. provided the in-house resources for the study, supervised the overall project, and provided domain knowledge expertise. All authors revised and commented on the manuscript.

\section*{Data availability}
The data used in the study are the proprietary of Bayer AG, and not publicly available.

\section{Acknowledgements}
A.Z. thanks Robin Williams and Nikki Hayward from Bayer\texttrademark\ Medical Information for providing expert insightful and in-depth feedback on the results of topic discovery.

%\bibliography{coling2020}

\end{document}